\documentclass[a4paper,twoside]{article}
\usepackage{cite}
\usepackage{graphicx}
\usepackage{psfrag}
\usepackage{subfigure}
\usepackage{url}
\usepackage{amsmath}
\usepackage{array}
\newcommand*{\degree}{$^\circ$}

\begin{document}
\title{The Parameter-Less Self-Organizing Map algorithm}

\author{Erik~Berglund,
 and~Joaquin~Sitte
\thanks {E.~Berglund is with the Division of Complex and Intelligent Systems,
Information Technology and Electrical Engineering,
University of Queensland,
        St. Lucia, QLD 4072, Australia (berglund@itee.uq.edu.au).}
\thanks {J.~Sitte is with the Smart~Devices~Lab, Queensland~University~of~Technology,
        Brisbane, QLD 4001 Australia (j.sitte@qut.edu.au).}}
\markboth{}{Berglund
\MakeLowercase{\textit{et al.}}: The Parameter-Less Self-Organizing
Map algorithm}

\maketitle

\begin{abstract}
The Parameter-Less Self-Organizing Map (PLSOM) is a new neural
network algorithm based on the Self-Organizing Map (SOM). It
eliminates the need for a learning rate and annealing schemes for
learning rate and neighbourhood size. We discuss the relative
performance of the PLSOM and the SOM and demonstrate some tasks in
which the SOM fails but the PLSOM performs satisfactory. Finally we
discuss some example applications of the PLSOM and present a proof
of ordering under certain limited conditions.
\end{abstract}

\section{Introduction}

The SOM~\cite{kohonen:1990a,ritter:1992} is an
algorithm for mapping from one (usually high-dimensional) space to
another (usually low-dimensional) space. The SOM learns the correct
mapping independent of operator supervision or reward functions that
are seen in many other neural network algorithms, e.g.
backpropagation perceptron networks. Unfortunately this unsupervised
learning is dependent on two annealing schemes, one for the learning
rate and one for the neighbourhood size. There is no firm
theoretical basis for determining the correct type and parameters
for these annealing schemes, so they must often be determined
empirically. The Generative Topographic Mapping (GTM)
\cite{gtm,gtm2,vellido:2003} is one attempt at addressing this.
Furthermore, since these annealing schemes are time-dependent, they
prevent the SOM from assimilating new information once the training
is complete. While this is sometimes a desirable trait, it is not in
tune with what we know of the adaptive capabilities of the organic
sensomotor maps which inspired the SOM~\cite{neuralplast}. There
have been several attempts at providing a better scaling method for
learning rate and/or neighbourhood size as well as taking some of
the guesswork out of the parameter estimation.
\subsection{Previous works} One
such attempt was done by G\"{o}ppert and Rosenstiel
\cite{goppert:1996}, where the SOM is used to approximate a
function, and the approximation is used as a neighbourhood size
decay parameter on a per-node basis. Unfortunately this is not
applicable to cases other than function approximation and it
requires knowledge of the desired approximation values of the
function, thus losing the major advantage of the SOM; unsupervised
learning. A more closely related approach would be the Plastic Self
Organising Map (PSOM) \cite{lang:2002}, where the Euclidean distance
from the input to the weight vector of the winning node is used to
determine whether to add new nodes to the map. This is similar to
the Growing Neural Gas (GNG) algorithms
\cite{fritzke:1994,fritzke:1995}, but maintains plasticity. Another
approach is the Self Organizing with Adaptive Neighbourhood Neural
Network (SOAN) \cite{iglesias:1999} which calculates the
neighbourhood size in the input space instead of the output space
like the SOM variants. The SOAN tracks the accumulated error of each
node, and scales the neighbourhood function accordingly between a
minimum and a maximum value, and like the GNG algorithms it can
increase or decrease the number of nodes. This still leaves several
parameters to be determined empirically by the user. The Time
Adaptive Self-Organizing Map (TASOM)
\cite{shah-hosseini:2000,shah-hosseini:2003} addresses the inability
of the SOM to maintain plasticity by keeping track of dynamic
learning rates and neighbourhood sizes for each individual node. The
neighbourhood size is dependent on the average distance between the
weight vector of the winning node $c$ and its neighbours, while the
learning rate is dependent only on the distance between the weight
of a given node $i$ and the input, similar to the standard SOM. The
user is still required to select several training parameters without
firm theoretical basis.  The Auto-SOM \cite{haese:1999,haese:2001}
uses Kalman filters to guide the weight vectors towards the centre
of their respective Voronoi cells in input space. This automates
computation of learning rates and neighbourhood sizes, but the user
is still required to set the initial parameters of the Kalman
filters. Unfortunately it is more computationally expensive than the
SOM, and this problem increases with input size and number of inputs
in the training set. The Auto-SOM also needs to keep track of all
previous inputs, which makes continuous learning difficult and
increases computational load, or compute the Voronoi set for each
iteration, which would increase computational load and is only
feasible if the input probability density distribution is known.
Other recent developments in self-organisation include the
Self-Organizing Learning Array \cite{starzyk:2005} and Noisy
Self-Organizing Neural Networks \cite{kwok:2004}.

\subsection{Overview}For these reasons we introduce the Parameter-Less
Self-Organizing Map (PLSOM). The fundamental difference between the
PLSOM~\cite{berglund:2003} and the SOM is that while the SOM depends
on the learning rate and neighbourhood size to decrease over time,
e.g. as a function of the number of iterations of the learning
algorithm, the PLSOM calculates these values based on the local
quadratic fitting error of the map to the input space. This allows
the map to make large adjustments in response to unfamiliar inputs,
i.e. inputs that are not well mapped, while not making large changes
in response to inputs it is already well adjusted to. The fitting
error is based on the normalised distance from the input to the
weight vector of the winning node in input space. This value
(referred to as $\epsilon$, the lowercase Greek letter epsilon,
throughout this paper) is computed in any case, hence this mechanism
can be implemented without inducing noteworthy increases in the
computational load of the map or hindering parallelised
implementations \cite{campbell:2005}. In Section \ref{ordinarysom}
we give details of the PLSOM algorithm, in Section \ref{performance}
we evaluate its performance relative to the SOM, in Section
\ref{analysis} we explain the observed behaviour of the PLSOM, in
Section \ref{applications} we give examples of applications, a brief
discussion of some aspects of the PLSOM relative to non-linear
mapping problems in Section \ref{discussion} and Section
\ref{conclusion} is the conclusion. For mathematical proofs, see
Appendix \ref{appendix1}.

\section{Algorithm}\label{ordinarysom}
As an introduction and to give a background for the PLSOM we will
here give a brief description of the SOM algorithm before we move on
to the PLSOM itself.
\subsection{The ordinary SOM algorithm}\label{stdsom}
The SOM variant we will examine is the Gaussian-neighbourhood,
Euclidean distance, rectangular topology SOM, given by Equations
(\ref{stdlrn})-(\ref{shrinknh}). The algorithm is, in brief, as
follows: An input $x(t)$ is presented to the network at time (or
timestep, iteration) $t$. The 'winning node' $c(t)$, i.e. the node
with the weight vector that most closely matches the input at time
$t$, is selected using Equation (\ref{winner}).
\begin{equation}\label{winner}
    c(t) = \arg \min_{i}(||x(t) - w_i(t)||_2)
\end{equation}
$w_i(t)$ is the weight vector of node $i$ at time $t$. $||.||_2$
denotes the $L^2$-norm or n-dimensional Euclidian distance. (The SOM
can use other distance measures, e.g. Manhattan distance.) The
weights of all nodes are then updated using Equations
(\ref{stdlrn})-(\ref{nhfunc}).

\begin{equation}\label{stdlrn}
    w_{i}(t+1) = w_{i}(t)+\Delta w_{i}(t)
\end{equation}
\begin{equation}\label{stdlrn1}
    \Delta w_{i}(t) = \alpha(t)h_{c,i}(t)[x(t)-w_{i}(t)]
\end{equation}
\begin{equation}\label{nhfunc}
    h_{c,i}(t) = e^{\frac{-d(i,c)^2}{\beta(t)^2}}
\end{equation}

$h_{c,i}(t)$ is referred to as the neighbourhood function, and is a
scaling function centred on the winning node $c$ decreasing in all
directions from it. $d(i,c)$ is the Euclidean distance from node $i$
to the winning node $c$ in the node grid. As is the case with the
input/weight distance, the node distance can be calculated using
some other distance measure than the Euclidean distance, e.g. the
Manhattan distance or the link distance, and the grid need not be
rectangular. $\alpha(t)$ is the learning rate at time $t$,
$\beta(t)$ is the neighbourhood size at time $t$.

Lastly the learning rate ($\alpha$) and neighbourhood size ($\beta$)
are decreased in accordance with the annealing scheme. One possible
annealing scheme is given by Equations (\ref{shrinkfunc}) and
(\ref{shrinknh}) for the decrease of the learning rate and the
neighbourhood size, respectively - the important point is that the
annealing scheme relies on the time step number $t$ and not the
actual fitness of the network.

\begin{equation}\label{shrinkfunc}
    \alpha(t+1) = \alpha(t)\delta_{\alpha},\hspace{6mm} 0 < \delta_{\alpha} < 1
\end{equation}
\begin{equation}\label{shrinknh}
    \beta(t+1) = \beta(t)\delta_{\beta}, \hspace{6mm} 0 < \delta_{\beta} < 1
\end{equation}

Here $\delta_{\beta}$ and $\delta_{\alpha}$ are scaling constants
determined beforehand.

These steps are repeated until some preset condition is met, usually
after a given number of iterations or when some measurement of error
reaches a certain level. The density of the nodes in input space are
proportional to the density of input samples, however this may lead
to undesired results, see Figure \ref{figure2}. Several variations
of the algorithm outlined here exists, e.g. the Matlab
implementation of the SOM uses a two-phased learning algorithm (an
ordering phase and a tuning phase) and a step-based neighbourhood
function.

\subsection{The PLSOM algorithm}\label{epsilon} The fundamental idea of the PLSOM is
that amplitude and extent of weight updates are not dependent on the
iteration number, but on how well the PLSOM fits the input data. To
determine how good the fit is, we calculate a scaling variable which
is then used to scale the weight update.
 The scaling variable, $\epsilon$, is defined in Equations (\ref{leasterr})
and (\ref{averageErr}).
\begin{equation}\label{leasterr}
    \epsilon(t) = \frac{||x(t) - w_c(t)||_2}{r(t)}\\
\end{equation}
\begin{equation}\label{averageErr}
 \begin{array}{ll}
    r(t) = \max(||x(t) - w_c(t)||_2,r(t-1)),\\
    r(0) = ||x(0) - w_c(0)||_2
    \end{array}
\end{equation}
$\epsilon(t)$ is best understood as the normalised Euclidean
distance from the input vector at time $t$ to the closest weight
vector. If this variable is large, the network fits the input data
poorly, and needs a large readjustment. Conversely, if $\epsilon$ is
small, the fit is likely to already be satisfactory for that input
and no large update is necessary.

\begin{figure}
 \centering
    \includegraphics[width= 200pt]{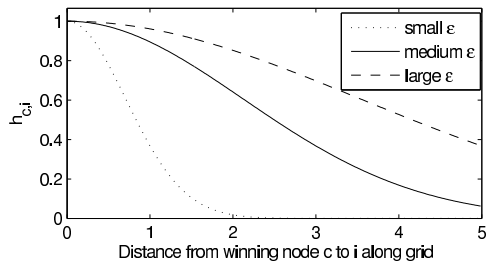}\\
    \caption{Plot showing the effect of different $\epsilon$ values on the neighbourhood function.}
    \label{hciplot}
\end{figure}

The algorithm for the PLSOM uses a neighbourhood size determined by
$\epsilon$, thus replacing the equation governing the annealing of
the neighbourhood with $\beta(t) = constant\hspace{2mm} \forall t$.
$\beta$ is scaled by $\epsilon(t)$ in the manner of (\ref{seuq}),
giving $\Theta(\epsilon(t))$, the scaling variable for the
neighbourhood function, Equation (\ref{nunhfunc}).
\begin{equation}\label{seuq}
    \Theta(\epsilon(t)) = \beta\epsilon(t), \hspace{6mm}\Theta(\epsilon(t)) \geq \theta_{min}
\end{equation}
(\ref{seuq}) is not the only option for calculating $\Theta$,
another example is (\ref{theta2}).
\begin{equation}\label{theta2}
    \Theta(\epsilon(t)) = (\beta-\theta_{min})\epsilon(t)+\theta_{min}
\end{equation}A third alternative is
(\ref{theta3}), which is used in generating Figures
\ref{sequence1}-\ref{sequence4}.
\begin{equation}\label{theta3}
    \Theta(\epsilon(t)) = (\beta-\theta_{min})ln(1 + \epsilon(t) (e-1))+\theta_{min}
\end{equation}
where $ln()$ is the natural logarithm, $e$ is the Euler number and
$\theta_{min}$ is some constant, usually 0 for Equation
(\ref{theta3}) or 1 for Equations (\ref{seuq})-(\ref{theta2}).
Equation (\ref{nunhfunc}) is the neighbourhood function.
\begin{equation}\label{nunhfunc}
    h_{c,i}(t) = e^{\frac{-d(i,c)^2}{\Theta(\epsilon(t))^2}}
\end{equation}
As before, $d(i,c)$ is a distance measure along the grid, i.e. in
output space, from the winning node $c$ to $i$ which is the node we
are currently updating. This gives a value that decreases the
further we get from $c$, and the rate of decrease is determined by
$\epsilon$, as can be seen in Figure \ref{hciplot}.
The weight update functions are Equations \ref{nolrnrt1} and
\ref{nolrnrt2}:
\begin{equation}\label{nolrnrt1}
    w_{i}(t+1) = w_{i}(t)+\Delta w_{i}(t)
\end{equation}
\begin{equation}\label{nolrnrt2}
    \Delta w_{i}(t) = \epsilon(t)h_{c,i}(t)[x(t)-w_{i}(t)]
\end{equation}

As we can see from Equation (\ref{nolrnrt2}) the learning rate
$\alpha(t)$ is now completely eliminated, replaced by $\epsilon(t)$.
Thus the size of the update is not dependent on the iteration
number. The only variable affecting the weight update which is
carried over between iterations is the scaling variable $r(t)$.
Practical experiments indicate that $r$ reaches it maximum value
after the first few iterations, and does not change thereafter.

\section{Performance}\label{performance}
The PLSOM completely eliminates the selection of the learning rate,
the annealing rate and annealing scheme of the learning rate and the
neighbourhood size, which have been an inconvenience in applying
SOMs. It also markedly decreases the number of iterations required
to get a stable and ordered map. The PLSOM also covers a greater
area of the input space, leaving a smaller gap along the edges.
\subsubsection{Comparison to the SOM variants}We trained the Matlab
SOM variant, the SOM and the PLSOM with identical input data, for
the same number of iterations. The input data was pseudo-random, 2
dimensional and in the $[0,1]$ range. This was chosen because a good
pseudo-random number generator was readily available, eliminating
the need to store the training data. Since the training data is
uniformly distributed in the input space the perfect distribution of
weight vectors would be an evenly spaced grid, with a narrow margin
along the edges of the input space. That way, each weight vector
would map an evenly sized area of the input space.

In comparing the two SOM implementations we used 3 separate quality
measures, which are all based on the shape and size of the cells. A
cell is the area in the input space spanned by the weight vectors of
four neighbouring nodes.
\begin{description}
\item[Unused space]~\\We summarised the area covered by all the
cells, and subtracted this from the total area of the input space.
The resulting graph clearly shows how the PLSOM spans a large part
of the input space after only a small number of iterations and
maintains the lead throughout the simulation (Figure
\ref{unspace1}). Please note that this quality measure will be
misleading in situations where cells are overlapping, but this will
typically only occur in the first few thousand iterations.

\begin{figure}
 \centering
    \includegraphics[height= 200pt,angle=-90]{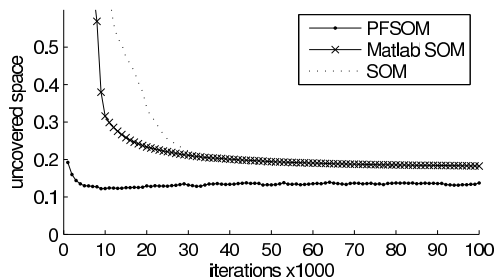}\\
    \caption{Graph of the decrease of uncovered space as training progresses for the PLSOM, the SOM and the Matlab SOM implementation. Note the quick expansion of the PLSOM and that it consistently covers a larger area than the SOM variants.}\label{unspace1}
\end{figure}

\item[Average skew]~\\For each cell we calculate the length of the
two diagonals in a cell and divide the bigger by the smaller and
subtract one, thus getting a number from 0 to infinity, where 0
represents a perfectly square cell. Again, we see that the PLSOM
outperforms the SOM in the early stages of simulation but after ca.
24000 iterations the SOM surpass the PLSOM. After 100000 iterations
the difference is still small, however. See Figure \ref{skew}.
\begin{figure}
 \centering
    \includegraphics[height= 200pt, angle=-90]{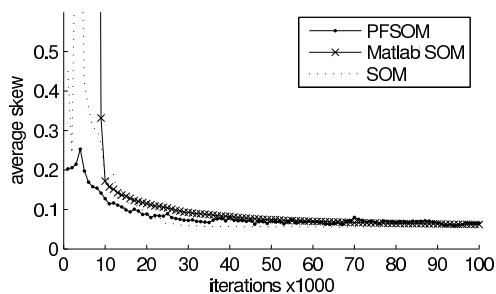}\\
    \caption{Graph of the average skew for the PLSOM, the SOM and the Matlab SOM implementation. For the first 24000 iterations the PLSOM is more ordered, before the SOM variants narrowly overtake it.}\label{skew}
\end{figure}
 \item[Deviation of cell size]~\\We calculate the
absolute mean deviation of the cell size and divide it by the
average cell size to get an idea of how much the cells differ in
relative size. Here the SOM is superior to the PLSOM after ca. 10000
iterations, mainly because of the flattened edge cells of the PLSOM,
see Figure \ref{meandev}. If we ignore the cells along the edge, the
picture is quite different: the PLSOM outperforms the SOM with a
narrow margin, see Figure \ref{middlemeandev}.
\end{description}

\begin{figure}
 \centering
    \includegraphics[height= 200pt, angle=-90]{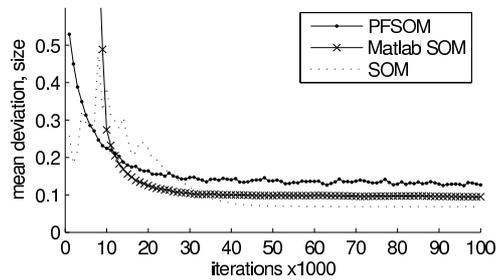}\\
    \caption{Graph of the absolute mean deviation of cell size for the PLSOM, the SOM and the Matlab SOM. The PLSOM is more regular up until ca. iteration 10000.}
    \label{meandev}
\end{figure}
\begin{figure}
 \centering
    \includegraphics[height= 200pt, angle=-90]{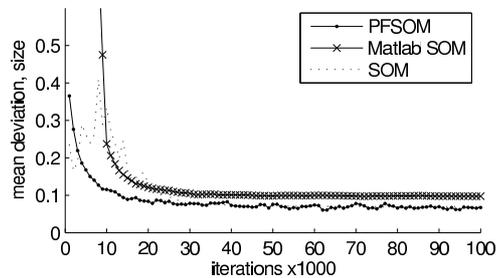}\\
    \caption{Graph of the absolute mean deviation of cell size for the PLSOM, the SOM and the Matlab SOM, excluding the edge cells. Compare to Figure \ref{meandev}. The PLSOM outperforms the Matlab SOM in both
adaptation time and accuracy, and the SOM needs until ca. iteration
30000 to reach the same level of ordering.} \label{middlemeandev}
\end{figure}

\subsubsection{Plasticity preservation}\label{plaspres}
\begin{figure}
 \centering
    \includegraphics[width= 130pt]{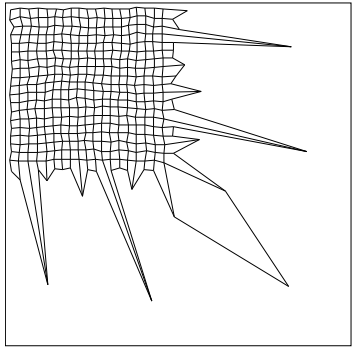}\\
    \caption{SOM first trained with inputs ranging from 0 to 0.5 for 50000 iterations shown after 20000 further training iterations with inputs ranging from 0 to 1.0.}
    \label{smallbig2}
\end{figure}

\begin{figure}
 \centering
    \includegraphics[width= 130pt]{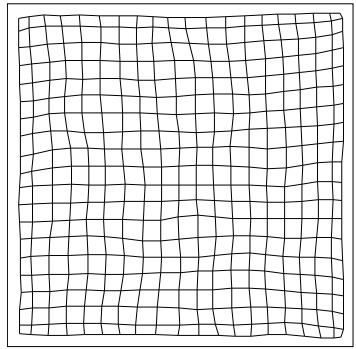}\\
    \caption{PLSOM first trained with inputs ranging from 0 to 0.5 for 50000 iterations shown after 20000 further training iterations with inputs ranging from 0 to 1.0. Note the difference between this and Figure \ref{smallbig2}.}
    \label{smallbig4}
\end{figure}

The illustrations in this
section show the positions of the weight vectors, connected with
lines, in the input space. When a SOM has been trained, it will not
adapt well to new data outside the range of the training data, even
if a small residual learning rate is left. This is illustrated by
Figure \ref{smallbig2}, where a SOM has been presented with
pseudo-random, uniformly distributed 2-dimensional data vectors in
the $[0, 0.5]$ range for 50000 iterations. Thereafter the SOM was
presented with 20000 pseudo-random, uniformly distributed
2-dimensional data vectors in the $[0, 1]$ range, after which the
SOM has adapted very little to the new data. In addition the
adaptation is uneven, creating huge differences in cell size and
distorting the space spanned by the weight vectors. If we subject a
PLSOM to the same changes in input range, the difference is quite
dramatic; it adapts correctly to the new input range almost
immediately, as seen in Figure \ref{smallbig4}.


\subsubsection{Memory}
\begin{figure}
     \centering
    \includegraphics[width= 130pt]{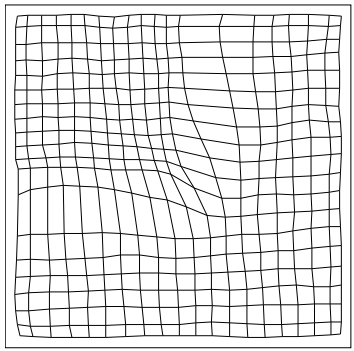}\\
    \caption{PLSOM first trained with inputs ranging from 0 to 1 for 50000 iterations shown after 20000 further training iterations with inputs ranging from 0 to 0.5.
    Note that while the weights have a higher density in the new input space, the same area as before is still covered, i.e. none of the old input space has been left uncovered.}
    \label{bigsmall4}
\end{figure}

In the opposite case, viz. the SOM is
presented with a sequence of inputs that are all restricted to a
small area of the training input space, it would be preferable if
the SOM maintains its original weight vector space, in order to not
'forget' already learned data. Figure \ref{bigsmall4} demonstrates
what happens to a PLSOM if it is trained with pseudo-random,
uniformly distributed 2-dimensional data in the $[0, 1]$ range for
50000 iterations and then presented with inputs confined to the $[0,
0.5]$ range for 20000 iterations. This leads to an increase of the
density of weight vectors in the new input space, yet maintains
coverage of the entire initial input space, resulting in distortions
along the edge of the new input space. Both these effects are most
pronounced in the PLSOM.

\subsection{Drawbacks}The PLSOM is measurably less ordered than a
properly tuned SOM and the edge shrinking is also more marked in the
PLSOM. The PLSOM does not converge in the same manner as the SOM
(there is always a small amount of movement), although this can be
circumvented by not performing new weight updates after a
satisfactory fit has been established.

\section{Analysis}\label{analysis}
This section highlights a special case where the SOM fails but the
PLSOM succeeds, and explores the causes of this.
\subsection{Experiments}
\begin{figure}
 \centering
    \includegraphics[width=130pt]{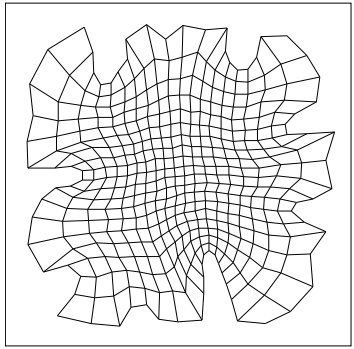}\\
    \caption{Ordinary SOM after 100000 iterations of normally distributed input with mean 0.5, standard deviation 0.2, clipped to the $[0,1]$ interval.
    Note that two nodes which are close in input space may not be close on the map.}
    \label{figure2}
\end{figure}

\begin{figure}
 \centering
    \includegraphics[width=130pt]{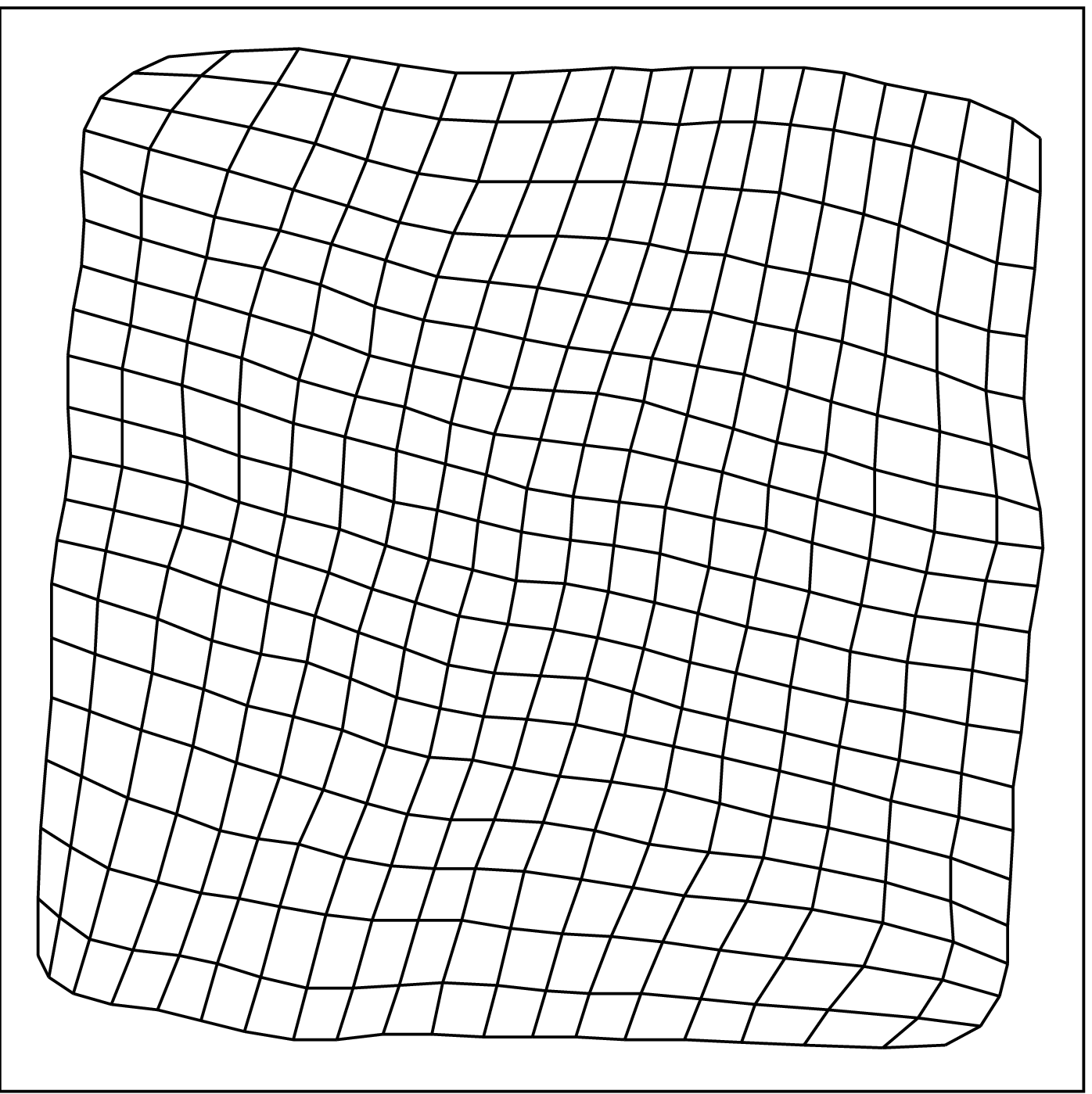}\\
    \caption{PLSOM after 100000 iterations of normally distributed input with mean 0.5, standard deviation 0.2, clipped to the $[0,1]$ interval.
    While the correspondence between weight vector density and input density is weaker than for the SOM, the topology is preserved.
    Compare to Figure \ref{figure2}. See also Figure \ref{densityspread}}
    \label{figure3}
\end{figure}

We have applied the PLSOM and two variants of the SOM to the same
problem; mapping a non-uniformly distributed input space. As input
space we used a normal distributed pseudo-random function with a
mean of 0.5 and standard deviation of 0.2. Values below 0 or above 1
were discarded. The same random seed was used for all experiments
and for initialising weights. A SOM variant that uses the same
neighbourhood function as the PLSOM and an exponential annealing
scheme for learning rate and neighbourhood size, here nominated
'plain SOM', was used for comparison. As can be seen from Figure
\ref{figure2} the SOM is severely twisted when we try a 20-by-20
node rectangular grid.
The size of the ordinary SOM algorithm must be reduced to 7-by-7
before all traces of this twisting are removed. Altering the
annealing time does not solve the problem. The PLSOM on the other
hand performs well with the initial size of 20-by-20 nodes, filling
the input space to over 77\%, see Figure \ref{figure3}.

\subsection{Explanation}
This phenomenon can be explained by looking at the likelihood of a
given input in relation to the size of the weight update this input
will result in, i.e. the expected update given the input
distribution. The likelihood of an input occurring is governed by
the Gaussian probability density function. The likelihood $p$ of an
input occurring in the interval $<z_1,z_2>$, where $z_1 \leq z_2$,
is approximated using the error function, $er\!\!f$:

\begin{equation}\label{erf1}
    p\,(z_1,z_2) = \frac{1}{2}\,(er\!\!f(\frac{z_2\,s}{\sqrt{2}}) - er\!\!f(\frac{z_1\,s}{\sqrt{2}}))
\end{equation}

where $s$ is a scaling constant to account for the standard
deviation. An analysis of the expected update of a given node is
given by Equation (\ref{step1}):
\begin{equation}\label{step1}
    \xi(\Delta w, x) = \Delta w(x) \rho(x)
\end{equation}

Where $\xi(\Delta w, x)$ is the expected displacement of weight
vector $w$ given $x$ as input, $\Delta w(x)$ is the displacement of
$w$ given $x$ and $\rho(x)$ is the probability density of this
input.

\begin{figure}
 \centering
    \includegraphics[width=200pt]{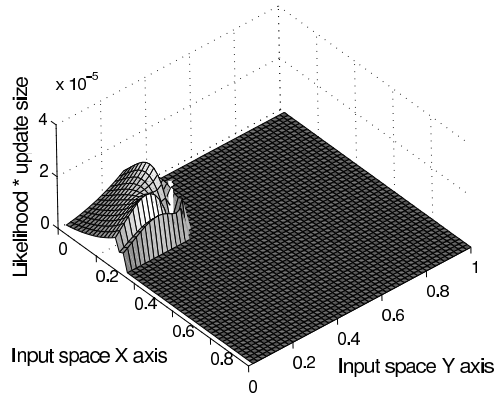}\\
    \caption{Update size x likelihood for a corner node $v$ of a 20x20 node ordinary SOM algorithm. The position of $v$ in the input space is marked by a vertical white line.
    The position of $v$ in the map is (1,1).}
    \label{somcrn}
\end{figure}

\begin{figure}
 \centering
    \includegraphics[width=200pt]{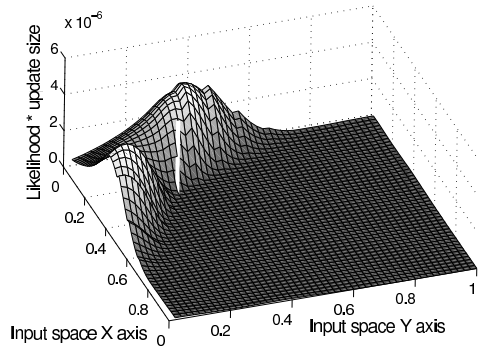}\\
    \caption{Update size x likelihood for a corner node $v$ of a 20x20 node PLSOM. The position of $v$ in the input space is marked by a vertical white line.
    The position of $v$ in the map is (1,1).}
    \label{plsomcrn}
\end{figure}

By discretising this over a 2-dimensional n-by-n grid, $N$,  we can
plot an approximation of the expected displacement for each square
of $N$, as seen in Figures \ref{somcrn}-\ref{plsomcrn}.

When comparing the expected update of a Matlab SOM algorithm and a
PLSOM we see that the PLSOM edge nodes receives a far larger amount
of its update from outside the area covered by the map than its
Matlab counterpart, thus making sure that the expansion outwards is
even and less jerky.


To get a clearer picture, we need to integrate the expected
displacement over the entire input space $\Omega$ which contains all
possible inputs $x$, giving Equation (\ref{step2}):
\begin{equation}\label{step2}
\xi(\Delta w) = \frac{1}{\Omega}\int_{\Omega}\Delta w(x)\rho(x) dx
\end{equation}

\begin{figure}
 \centering
    \includegraphics[width=150pt]{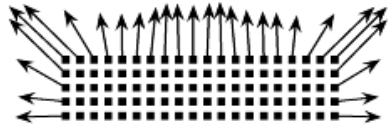}\\
    \caption{The expected displacement vectors for the edge nodes along one edge of an ordinary SOM.
    Note that the vectors are changing direction abruptly from node to node, causing the warping.}
    \label{msomv}
\end{figure}

\begin{figure}
 \centering
    \includegraphics[width=150pt]{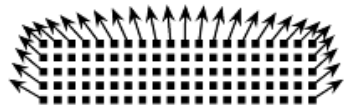}\\
    \caption{The expected displacement vectors for the edge nodes along one edge of a PLSOM.}
    \label{plsomv}
\end{figure}

Discretising the integrated expected displacement gives a vector for
each node in the map, indicating how much and in which direction it
is likely to be updated given the input distribution, as shown in
Figures \ref{msomv} and \ref{plsomv}.

As we can see from Figure \ref{msomv}, this vector is greater for
the corner node than for the side node in the SOM algorithm, while
the opposite is true for the PLSOM, as seen in Figure \ref{plsomv}.
This leads the corner nodes in the SOM algorithm to expand outwards
faster than the side nodes, thus creating the warping. In the PLSOM,
the side nodes expand outward faster, creating an initial 'rounded'
distribution of the weights, but subsequent inputs pull the corners
out. Also note that the edge nodes of the PLSOM is only marginally
pulled inwards by inputs inside the weight grid, since the amount of
update depends on the distance from the input to the closest node,
not only on the distance from the node in question - this
contributes to the quicker, more even expansion.

Finally, the weight update functions of the different algorithms
give us the last piece of the explanation. Consider a map that
receives an input far outside the area it is currently mapping,
after already being partly through its annealing, and therefore
partially ordered.

When this happens to an ordinary SOM the update will be large for
the winning node, but because the size of the neighbourhood function
is so small the neighbours of the winning node will receive only a
very small update. If the same situation occurs to a PLSOM, the
neighbourhood size will scale (since it is dependent on the distance
from the input to the winning node) to include a larger update for
the neighbours of the winning node, thus distributing the update
along a larger number of the edge nodes.

\begin{figure}
  \centering
  \includegraphics[width=230pt]{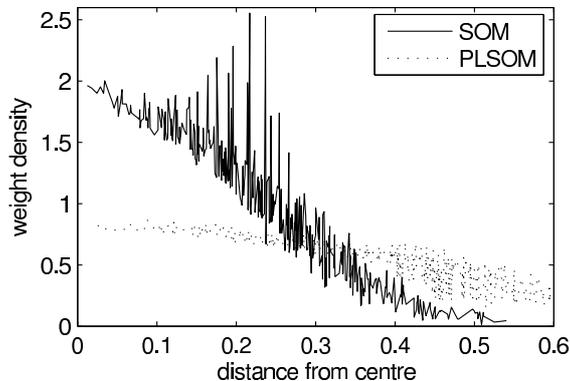}\\
  \caption{Weight density vs. distance from centre for the SOM and the PLSOM.
  The 2-dimensional input was normal distributed with a 0 mean and 0.2 standard deviation.
  Observe that while the PLSOM has less correlation between input density and weight density,
  it has far less variance and covers a larger area. See also Figures \ref{figure2} and \ref{figure3}}
  \label{densityspread}
\end{figure}

It should be pointed out that mapping a large portion of the input
space while preserving such a skewed distribution is not possible -
the difference between the length along the edges and the length
through the centre is too great to preserve neighbourhood relations.
When faced with this type of high-variance input distribution, one
is faced with the choice of which property to sacrifice;
neighbourhood consistency or density equivalence. The SOM tries to
do both, and fails. GNG, PSOM and similar algorithms do both at the
cost of ending up with arbitrary network connections. The PLSOM is
unique in preserving neighbourhood relations for a pre-defined
network. This comes at a cost of poorer correspondence between input
density and weight density, as can be seen from Figure
\ref{densityspread}.

\section{Applications}\label{applications}
The PLSOM has been applied to three familiar problems by the
authors. These applications will only be explored briefly here, in
the interest of not distracting from the main subject of this
article.

\subsection{Sound source localisation through active audition}
This application deals with processing a stereo sound signal,
presenting it to a PLSOM or SOM to determine the direction of the
sound source and orienting the microphones towards the sound source.
This application illustrates that the PLSOM can deal with cases
where the number of input dimensions (512) is far higher than the
number of output dimensions (2).

\subsubsection{Earlier works}
The physics of binaural audition is discussed in
\cite{king:1995,nandy:1995,hartmann:1999} and reinforcement learning
in \cite{sutton:1992}. Early works include
\cite{huang:1995,huang:1997}, but it is important do distinguish
between passively determining the direction of sound sources and
active audition. Active audition aims to using the movement of the
listening platform to pinpoint the location, in the same way
biological systems do. Active audition is therefore an intrinsically
robotic problem. Other active audition works can be grouped into
subcategories: First we have applications that rely on more than two
microphones to calculate the source of a sound, e.g.
\cite{rabinkin:1996}. While these are certainly effective, we
observe that in nature two sound receivers are sufficient. Since
microphones consume power and are a possible point of failure, we
see definite advantages to being as frugal as nature in this
respect. Secondly we have methods relying on synergy of visual and
auditory cues for direction detection, most notably by members of
the SIG Humanoid group \cite{nakatani:1994,
kitano:2002,nakadai:2000a,lourens:2000,nakadai02realtime,nakadai:2003}.
Some of these also include neural networks and even SOMs, such as
\cite{rucci:1999,nakashima:2002a}. However, it is known that even
humans that are born blind can accurately determine the direction of
sounds \cite{zwiers:2001}, so interaction between vision and hearing
cannot be a crucial component of learning direction detection.
Our implementation is unique in that it does not rely on visual
cues, specialised hardware nor any predefined acoustic model. It
learns both the direction detection and the correct motor action
through unsupervised learning and interaction with its environment.
It also incorporates a larger number of measures than other methods,
which is made possible through the SOM/PLSOM ability to find
patterns in high-dimensional data.

\subsubsection{System description} 
\begin{figure}
 \centering
 \includegraphics[width=200pt]{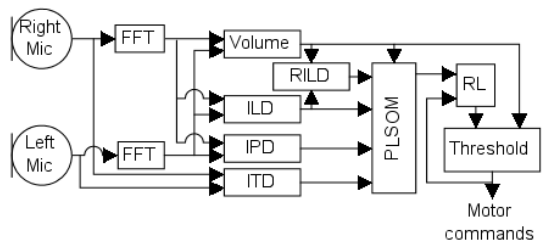}
 \caption{Active audition system layout}
 \label{soundflow}
\end{figure}

The aim is to let the process be
completely self-calibrating - all that is needed is to provide a set
of sound sources, and the algorithm will figure out on its own where
the sound is coming from and how to orient towards it. This is done
using a pipelined approach:
\begin{enumerate}
\item Digital stereo samples are streamed from a pair of
microphones.
\item For each sampling window, which is 512 samples long, we
compute the Fast Fourier Transform (FFT) of the signal. This is
averaged to 64 subbands.
\item For each sampling window we
compute Interaural Time Difference (ITD), Interaural Level
Difference (ILD), Interaural Phase Difference (IPD) and Relative
Interaural Level Difference (RILD). ILD, IPD and RILD are based on
the FFT, so that we have one value for each of the 64 subbands.
\item The resulting 256-element vector is presented to a PLSOM.
\item The position of the winning node is used as index into a
weight matrix which selects the appropriate motor action.
\item If the sound volume is over a given threshold, the selected
motor action is carried out.
\item After a short delay, the algorithm checks whether the winning
node has moved closer to the centre of the map, and uses this to
calculate a reward value for the reinforcement learning module.
\end{enumerate}
Figure \ref{soundflow} illustrates this procedure.
The pipelined approach has the advantage of making experimenting
with different processing paths much simpler. It also lends the
approach to parallelising hardware and software implementations. In
order to train the PLSOM a number of samples are recorded. We used
white noise samples from 38 locations in front of the robot. The
samples were recorded at 50 and 300 cm distance from the source,
with 10\degree ~horizontal spacing. The training algorithm then
presents a few seconds of random samples to the system for 10000
training steps. Each sample is 256-dimensional, and a new sample is
presented every 32 ms. This sensitises each node to sound from one
direction and distance. The latter part of the training is done
online, with the robot responding to actual sound from a stationary
speaker in front of it. Initially the robot head is pointed in a
random direction, and the training progresses until the robot keeps
its head steadily pointed towards the speaker, at which time a new
random direction is picked and the training continues.

\subsubsection{Results}
\begin{figure}
 \centering
 \includegraphics[width=200pt]{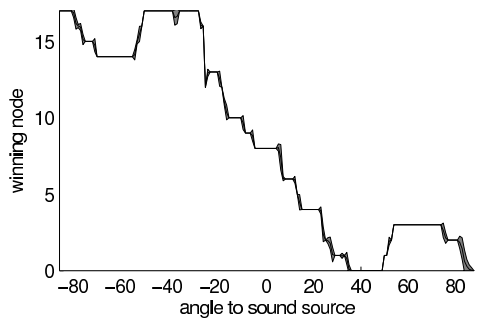}
 \caption{High dimensional data: Average winning node vs. actual angle using the PLSOM method. The grey area is one standard deviation.}
 \label{plot98}
\end{figure}

Our approach described above consistently manages an accuracy of
around 5\degree , which is comparable to human acuity. The precision
was determined by keeping the robot stationary and registering the
winning node of the PLSOM. We then moved the sound source
horizontally until the winning node stabilised on one of the
immediate neighbours of the initial winning cell, noting how much
the source had to be moved. The relative accuracy of the method is
demonstrated in Figure \ref{plot98}. The graphs were generated using
a set of recordings of white noise at a distance of 1 metre.
The PLSOM is, as we can see, almost free of deviation. This enables
one to estimate the direction using a small number of samples, i.e.
quickly.

\subsection{Inverse Kinematics}
Inverse Kinematics (IK) is the problem of determining the joint
parameters of a robotic limb for some given position. This problem
is interesting in evaluating the PLSOM because it involves mapping
between two spaces with wildly different topologies, a half-torus
shaped space in Euclidean space and a roughly wedge-shaped space in
joint parameter space.
\subsubsection{Existing methods}
Robot control depends on IK but there are several problems
associated with the existing methods. The Jacobian pseudo-inverse
lets one solve the problem completely, but is computationally
expensive, relatively complicated and unstable around singularities.
The Jacobian transpose is faster and simpler, but not particularly
accurate and does not move in a shortest path like the
pseudo-inverse. Other methods, like Cyclic Coordinate Descent
\cite{wang:1991}, which solve the unstable singularity problem has
been proposed. There have also been solutions of the IK problem
using SOMs \cite{ritter:1992}, which lets a SOM learn the joint
angles for a number of points in space, and an approximation of the
inverse Jacobian in the vicinity of each point. This gives good
results with relatively few nodes but it relies on information being
stored in the nodes of the SOM, rather than holistically in the
network, leading each node to be more complicated than necessary. It
is not clear whether using an approximated inverse Jacobian can lead
to the same sort of instability around singularities as with the
pseudo-inverse Jacobian. Even so, our method borrows heavily from
the SOM approach and should be seen in relation to it.
\subsubsection{Proposed solution}
We opted for a slightly different and, we believe, novel approach
using the PLSOM. Each node maps a point in 3D space to a point in
joint parameter space. The map is trained through generating a random
joint configuration and updating the PLSOM weights with it. After
the training is complete each node is labelled with the manipulator
position that will result from applying the weight vector to the
joint parameters. This allows the manipulator to be positioned in
accordance with the following simple algorithm:
\begin{enumerate}
\item Select the node closest to the desired point in space by comparing node labels. This
step can be greatly accelerated by starting the search at the last
node used.
\item Select the neighbours of the node so that we can create three
almost orthogonal vectors in 3D space. This is trivial given the
lattice structure of the PLSOM.
\item The 3 vectors in 3D space are then orthogonalised using
the Gram-Schmidt algorithm.
\item The analogous steps are carried out on the 3 corresponding vectors
in joint parameter space.
\item The resulting vectors can now be used to interpolate the joint
parameters.
\end{enumerate}

\subsubsection{Experiments and Results}
The PLSOM solves the IK problem very quickly, as no iteration is
necessary. One can input the desired target position and immediately
get an approximation of the joint parameters that achieve this
target. However, the level of precision depends on the number of
nodes in the network. As a demonstration of the capabilities of the
PLSOM method a 6 DOF robotic arm was programmed to play chess
against itself. A PLSOM with 3600 nodes was trained in a half-torus
shaped area in front of the robot covering a small chessboard and
the surrounding table, 30000 training iterations are completed in
less than 5 minutes on a low-end desktop PC. Even with the
relatively few nodes the error is well below the mechanical error in
the robotic arm. The robot is able to quickly and accurately pick
and place the chess pieces, as shown in the short video clip in
\cite{berglund:2004}.

\begin{figure}
 \centering
 \includegraphics[width=200pt]{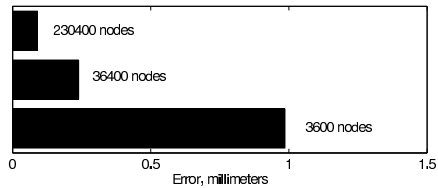}
 \caption{Error after 500 iterations of some different sized PLSOMs.}
 \label{errorplot}
\end{figure}

In order to assess the different IK methods, we performed a
simulation wherein a 3 DOF robot arm is moved from one position to
another. Each method is allowed 500 iterations to complete this and
for each iteration the error is calculated. We performed this test
with  a PLSOM with 3600 nodes, a PLSOM with 36400 nodes and a PLSOM
with 230400 nodes. We then repeated the experiment for 4 different
target positions and averaged the error. The result is displayed in
Figure \ref{errorplot}.
As we can see from the graph, the PLSOM's accuracy is related to the
number of nodes. It should be noted that training time is roughly
$O(n)$ in number of nodes $n$, so increasing accuracy is not
computationally expensive. Accuracy is slightly better than what is
reported in \cite{ritter:1992}, i.e. 0.02\% error compared to 0.06\%
error, although this is not surprising given the large difference in
number of nodes. The execution speed was measured and averaged over
all the experiments - the difference in execution speed for the 3600
node network and the 230400 node network is typically less than 2\%,
e.g. 29.52$\mu s$ and 29.98$\mu s$ on a low-end desktop pc.

\subsection{Classification of the ISOLET  data set.}
In order to test whether the PLSOM can handle very high-dimensional
and clustered data, we selected the ISOLET \cite{cole:1990} data set
for analysis. The data set contains suitably processed recordings of
150 speakers saying the name for each of the letters in the
alphabet, twice. The set is subdivided into the training set,
containing 120 speakers, and the test set with the remaining 30
speakers. Algorithms seeking to solve this problem are trained on
the training set and evaluated on the test set. The data is present
as 617-dimensional real-valued vectors with elements in the $[-1,
1]$ range representing various preprocessed properties of the sound,
see \cite{cole:1990} for details. The best result reported without
any further signal processing is 95.83\% \cite{dietterich:1991} and
95.9\% \cite{fanty:1990}, achieved with backpropagation. k-Nearest
Neighbour \cite{fix:1952} methods achieve from 88.58\% (k = 1) to
92.05\% (k = 5, ambiguities resolved by decreasing k) but are slow
due to the size of the training set which forces the computation of
6238 617-dimensional Euclidean distances for each classification.
Classification of clustered data is troublesome with the PLSOM
because it tries to approximate a low-dimensional manifold in the
input space, in this case there may not be a manifold. Since the
PLSOM is unsupervised there is no way to direct learning effort to
one particular property of the input, the PLSOM tries to map them
all. In the ISOLET set the letter is not the only property encoded
in the data, there are all sorts of possible data that may or may
not be present such as speaker, the speakers' age, gender, dialect,
smoker/non-smoker and so on. Ideally the PLSOM would have one output
dimension for each dimension of the embedded manifold, but this is
impractical in this case. We therefore settled on a 3-dimensional
PLSOM with 20x20x20 nodes. The PLSOM was trained with random samples
from the training set for 100000 iterations with neighbourhood size
= 2, then each node is labelled with the input it responds to. If a
node responds to more than 1 input, a vote is performed and the node
is labelled with the input it responds to most frequently. If no
input gets thrice as many votes as other inputs, or if the node does
not respond to any input, it is removed. The map is then used to
classify the test set. This typically results in ca. 2800 remaining
nodes, which we utilise as the reference vectors for k-NN
classification. Thus the PLSOM can be seen as a way of speeding up
the k-NN algorithm by reducing the number of reference vectors. This
does however come at a price of accuracy - it achieves 90.31\%
accuracy at k=5.

\section{Discussion}\label{discussion}
\begin{figure}%
\centering \subfigure[SOM.]{
          \label{skewsom}
          \includegraphics[width= 100pt]{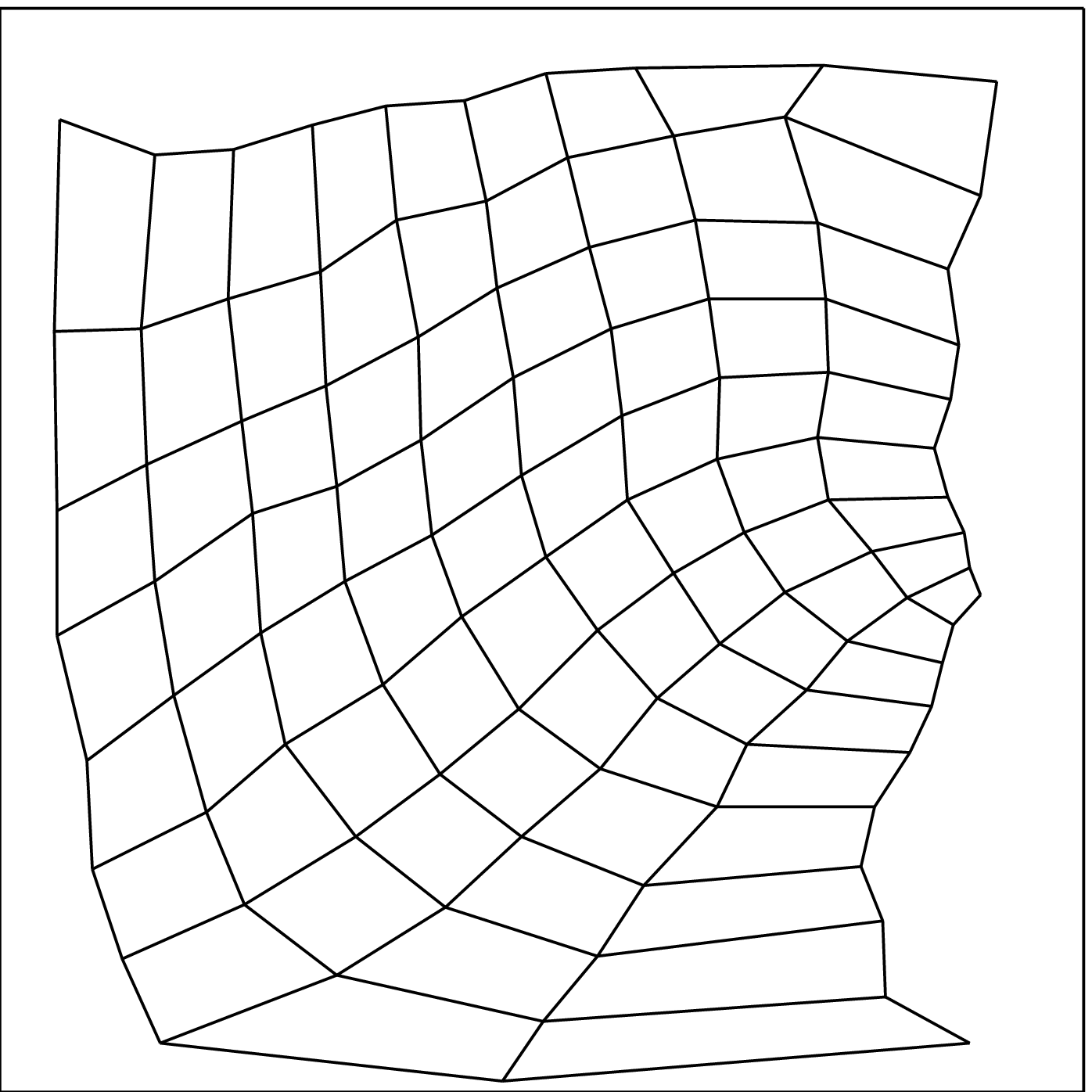}}%
          \hspace{6mm}
\subfigure[PLSOM.]{
          \label{straightplsom}
          \includegraphics[width= 100pt]{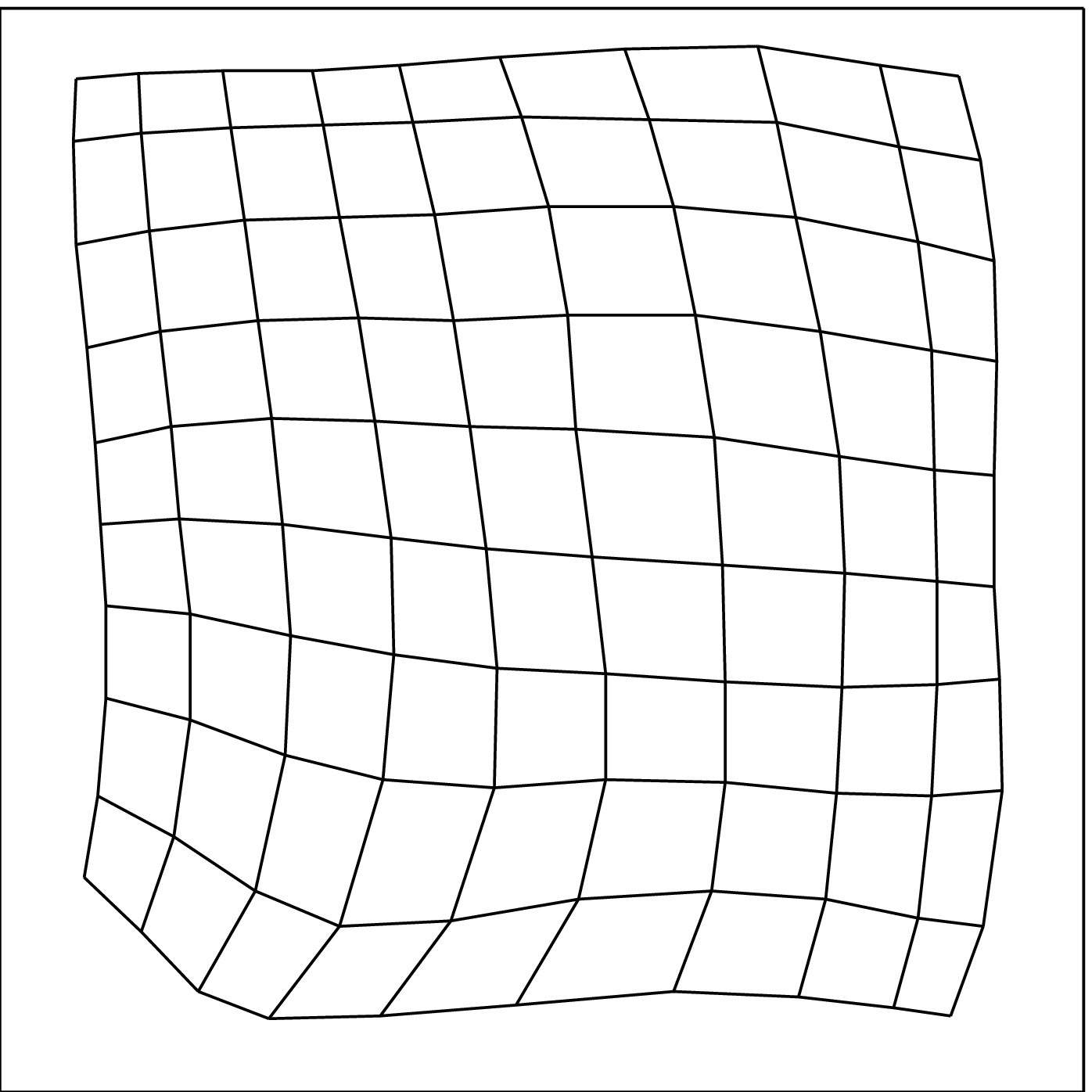}}\\
          \label{beacontest}
          \caption{The 3-beacon navigation mapping by the SOM and the PLSOM in the unit square.
          Both maps have a neighbourhood size of 17. Beacons were positioned at
          (0.3,-0.3),(1.3,0.5) and (-0.5,0.8). The origin is in the upper left-hand corner.}
\end{figure}

As indicated above, the PLSOM emphasises maintaining topology over
modelling the input density distribution. In many cases this may be
a disadvantage, for example with highly clustered data. In other
cases, however, this is exactly what is needed for a useful mapping
of the input space. This happens when the input space is nonlinearly
mapped to the space from which the inputs are selected. This is best
illustrated by an example (based on \cite{keeratipranon:2005}): A
robot navigates a square area by measuring the angles between three
uniquely identified beacons, giving a 3-dimensional vector with a
one-to-one relationship to the location of the robot in space.
Unfortunately the density and topology of the embedded manifold is
quite different from the input space, causing a large number of
samples to fall in a few small areas of the 3-dimensional input
space where the curvature of the embedded Riemannian manifold is
high. This causes the SOM to try and fit a corner of the map where
there should be no corner, see Figure \ref{skewsom}.
The density of the input space is correctly represented, but the
topology of the embedded manifold (which is the interesting property
in this case) is not. This can of course be corrected by selecting
the samples carefully in the case of our example since we know the
bidirectional mapping, but in a real application this would not be
known. It would therefore be beneficial to find an approach that
largely ignores the distribution of the input space and instead
emphasises the topology of the embedded manifold. The PLSOM does
this to a higher degree, see Figure \ref{straightplsom}.

\section{Conclusion}\label{conclusion}
We have addressed several problems with the popular SOM algorithm
and proposed an algorithm that solves these; the PLSOM which
provides a simplification of the overall application process, since
it eliminates the problems of finding a suitable learning rate and
annealing schemes. The PLSOM also reduces the training time.
Flexibility and ordering of the map is facilitated and we have shown
that the PLSOM can be successfully applied to a wide range of
problems. Furthermore, the PLSOM is able to handle input probability
distributions that lead to failure of the SOM, albeit this comes at
the cost of lower correspondence between the input distribution and
the weight density. While the PLSOM does not converge in the sense
that a SOM will if the learning rate is allowed to reach $0$, the
same effect can be achieved by simply not performing weight updates
after a number of inputs. All this is achieved without inducing a
significant computation time increase or memory overhead. Finally we
have shown (see appendix \ref{appendix1}) that the PLSOM is
guaranteed to achieve ordering under certain conditions.

\appendix
\section{Proof of guaranteed ordering of a PLSOM with 3
nodes and 1-dimensional input and output.}\label{appendix1} This
section will present a proof of guaranteed ordering in a special
case of the PLSOM. We start out by establishing some lemmas
necessary for the proof, then we examine the proof and finally we
speculate implications of the proof. For all these proofs we assume
that:
\begin{equation}\label{prem1}
\epsilon \leq 1
\end{equation}
Where $\epsilon$ is the normalised distance from  the input to the
weight of the winning (closest) node.
\begin{equation}\label{prem2}
h_{c,i} \leq 1
\end{equation}
Where $h_{c,i}$ is a neighbourhood function which depends on
$\epsilon$ and the distance in output space between the winning node
$c$ and the current node $i$. Please see Section \ref{ordinarysom}
for a discussion of the neighbourhood function. $c = i \Rightarrow
h_{c,i} = 1$ and $h_{c,i}$ is monotonously decreasing with
increasing distance from $c$.
\begin{equation}\label{prem3}
h_{c,n} < h_{c,n+1}
\end{equation}\\
Which implies that $w_{n+1}$ is closer to the input than $w_n$, and
will hence receive a larger scaling from the neighbourhood function.
In the following an ordered map will denote a map where all the
nodes are monotonously increasing or decreasing, which means either
\ref{ord1} or \ref{ord2} is true.
\begin{equation}\label{ord1}
w_n < w_{n+1}, \forall n
\end{equation}
\begin{equation}\label{ord2}
w_n > w_{n+1}, \forall n
\end{equation}
Any other map will be called unordered.\\ \\
\textbf{Lemma 1:}\label{lemma1} The weights of a node cannot overshoot the input, i.e. the weights of a node cannot move from one side of an input to the other as a result of that input.\\
\textbf{Proof:} Assume\footnote{Note that for this section we
disregard the $(t)$ part of the notation, as it is implied.} that
there is an input $x$, and a node $i$ with weight $w_i$. The amount
of update to $w_n$, $\Delta w_i$ is equal to $\epsilon h_{c,i}[x -
w_i]$ as before. Since $\epsilon  \leq 1$ and $h_{c,i} \leq 1$, it
is clear that $|\Delta w_i| \leq |x - w_i|$. Also, $|\Delta w_i| =
|x-w_i|$ is only true where $\epsilon =
1 \wedge h_{c,i} = 1$, which can only hold for the winning node $c$. This proof also applies to the standard SOM where $\alpha (t) \leq 1$.\\
\\
\textbf{Lemma 2:} There exists no input that can turn an ordered 1
input dimension and 1 output dimension
map into an unordered one.\\
\textbf{Proof:} Proof by contradiction. Assume that $w_n$  and
$w_{n+1}$ are weights of an ordered 1D map, and $w_n < w_{n+1}$. We
will prove that no input $x \geq w_{n+1}$ can move $w_n$ past
$w_{n+1}$ (It is easy to see that the converse has to be true for $x
\leq w_n$). For $w_n$ to move past $w_{n+1}$ and making the map
unordered, the update of node $n$ must be greater than or equal to
the distance between node $n$ and node $n+1$, plus the update to
node $n+1$, so (\ref{false1}) has to be true.
\begin{equation}\label{false1}
\Delta w_n \geq w_{n+1} - w_n + \Delta w_{n+1}
\end{equation}
Remember that $\Delta w_n = \epsilon h_{c,n}(x-w_n)$ and that $x
\geq w_{n+1}
 > w_n$, which gives us (\ref{false2})
\begin{equation}\label{false2}
\begin{array}{l}
\epsilon h_{c,n}(x-w_{n+1})+ \epsilon h_{c,n}(w_{n+1} - w_n)\\
\geq w_{n+1}-w_n+\epsilon h_{c,n+1} (x - w_{n+1})
\end{array}
\end{equation}
It is clear that, because of the premises, Equation (\ref{false2})
cannot be true. This is a restatement of the proof that the ordered
states are absorbing sets, see \cite{cottrell:1998}
\\ \\
That leaves the cases where $w_n < x < w_{n+1}$. Since, according to
lemma 1, only one node (the winning node) can reach $x$ and no node
can overshoot $x$, it follows that the nodes must be on the same
side of $x$ after the weight update as before. Therefore, they
cannot become unordered.

 Note that the lemma 1 and 2 holds for PLSOMs with any number of
 nodes, as long as there is one input and one output dimension.
This proof is very similar to the one given for the SOM by Kohonen \cite{kohonen:1998}.\\ \\
\textbf{Lemma 3:}\label{lemma3} In the special case where $w_i = a, \forall i$ and some value $a$, any input other than $a$ will result in an ordered map.\\
\textbf{Proof:} Since all nodes have the same distance to the input,
the winning node will automatically be the first one, $w_0$. Again,
since all nodes are the same distance from the input, the amount
each node is updated is determined solely on the lattice distance
from the winning node. Therefore, $w_0$ will move the most, $w_1$ a
little less and $w_2$ even less and so on - resulting in an ordered
map.\\ \\
\textbf{Lemma 4:}\label{lemma4} A 1-dimensional PLSOM with 3 nodes
will always reach an ordered state given a sufficiently large number
of uniformly
distributed inputs.\\
\textbf{Proof:} The proof is computer-assisted, here we give an outline of the procedure for calculating it.\\
In short the proof is as follows:
\begin{enumerate}
 \item Calculate a scalar field in
expressing how much closer the weights are to an attractor point in
ordered space after an update, where positive values indicate that
the weights have moved away from the attractor.
\item Calculate the gradient of the scalar field. \item Calculate
the upper bound of the gradient.

\item Given the upper bound of the gradient we calculate the size of
a sample grid.


\item Given the upper bound of the gradient and the expected update
at each sample point, the expected update must point towards the
attractor in the vicinity of the sample point. If this holds for all
sample points, it holds for the whole subspace of unordered weights.
\end{enumerate}

\begin{figure}
 \centering
    \includegraphics[width= 160pt]{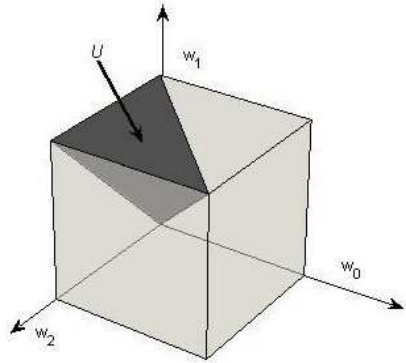}\\
    \caption{The unordered subspace $U$. All other unordered states are mirrors or inversions of states in this subspace.}
    \label{uspace}
\end{figure}

\begin{figure}
 \centering
    \includegraphics[width= 160pt]{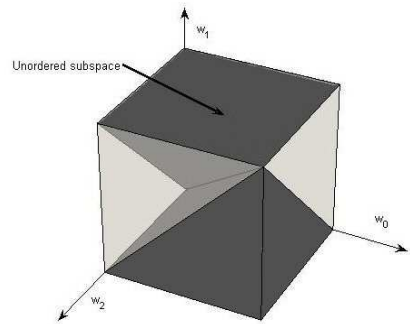}\\
    \caption{All unordered states in the volume of all possible states.}
    \label{unspace}
\end{figure}

\begin{figure}
 \centering
    \includegraphics[width= 160pt]{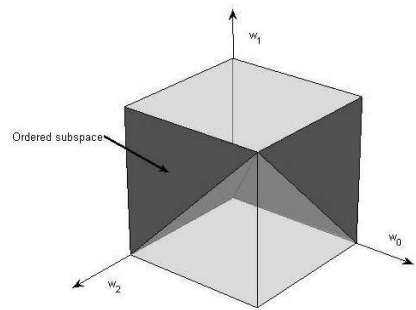}\\
    \caption{All ordered states in the volume of all possible states.}
    \label{ospace}
\end{figure}

The weights of the three PLSOM nodes are denoted $w_0,w_1,w_2$. The
conditions under which the proof is calculated are as follows:
\begin{itemize}
\item Uniformly distributed input in the range $[0,1]$.
\item $w_0 \leq w_2 \leq w_1$
\item $w_0 < w_1$
\item $0 \leq w_0 < 1$
\item $0 \leq w_2 \leq 1$
\item $0 < w_1 \leq 1$
\item Linear neighbourhood function, for simplicity.
\end{itemize}
The weights $w_0$, $w_1$ and $w_2$ can be seen as coordinates in a
3-dimensional space, where all possible configurations fill the unit
cube. The subspace spanned by the constraints above are denoted $U$,
see Figure \ref{uspace}. This subspace represents the only way in
which a 1D 3-node map can be unordered - all other unordered states
are inversions or mirrors of this state, see Figure \ref{unspace}.
An ordered map fulfils one of the following configurations:
\begin{enumerate}
\item $w_0<w_1<w_2$
\item $w_0>w_1>w_2$
\end{enumerate}
The ordered subspace fills the volume drawn in Figure \ref{ospace}.
Now on to the proof proper:\\
 We introduce the concept of the expected update
vector: Given a weight configuration and an input probability
density function we can compute the distance and direction that a
node is most likely to move in and is given in
(\ref{expectedupdate}).
\begin{equation}\label{expectedupdate}
\vec{u} = [u_0,u_1,u_2]
\end{equation}

Where $\vec{u}$ is the expected update vector, its elements given by
(\ref{uvec}).
\begin{equation}\label{uvec}
\begin{array}{r}
u_n = \frac{w_0+w_2}{2}\int_0^{\frac{w_0+w_2}{2}}f_n(0) dx\\
+\frac{w_1-w_0}{2}\int_{\frac{w_0+w_2}{2}}^{\frac{w_2+w_1}{2}}f_n(2)dx\\
+(1-\frac{w_2+w_1}{2})\int_{\frac{w_2+w_1}{2}}^{1}f_n(1) dx
\end{array}
\end{equation}

Where $f_n(c)$ is the expected update of node $n$ given $c$ as the
winning node. As mentioned above, we use a simplified version of
$f_n(c)$ to facilitate integration, see (\ref{fnc}).

\begin{equation}\label{fnc}
f_n(c) = \frac{|x-w_c|}{r}\left(1-\frac{|c-n|}{\beta}\right)(x-w_n)
\end{equation}

Where $r$ is the normalising variable and $\beta$ is the
neighbourhood size. For simplicity we set $r = 1$ and $\beta = 2$.
$x$ is the input, uniformly distributed in the $[0,1]$ interval.

As mentioned above, the three input weights of the nodes, $w_0,w_1$
and $w_2$ can be seen as coordinates in a 3-dimensional Euclidean
space, of which $U$ is a subspace.

Every point $\vec{w}$ in this subspace is associated with an
expected update $\vec{u}$: The position before an input is
represented by $\vec{w}$, and the most likely position after a
uniformly distributed random input is $\vec{w}+\vec{u}$.\\
Now we introduce a point denoted $\vec{t}$ in the \emph{ordered}
subspace of the unit cube, which is the attractor in the dynamic
system in $U$. In other words, all the expected weight updates in
$U$ will bring the weight vectors closer to the attractor, and hence
closer to the ordered subspace, see (\ref{difflen}).
\begin{equation}\label{difflen}
||\vec{w}+\vec{u}-\vec{t}||_1-||\vec{w}-\vec{t}||_1 < 0, w\in U
\end{equation}
where $||.||_1$ is the $L^1$-norm or Manhattan distance. The
$L^1$-norm was chosen because it produces a simpler expression than
the $L^2$-norm. $\vec{t}$ has been found empirically to be close to
$[\frac{377}{1000},\frac{121}{200},\frac{7}{10}]$, the exact
location is not important to this proof. Equation (\ref{difflen})
defines a 3-dimensional scalar field and in order to prove
negativity we compute the upper bound of the length of the gradient,
$16.5$. With this estimated upper bound we must check that no point
is further away from a sample point than $1.333*10^{-4}$, and that
no sample point has a value greater than $d = -2.2*10^{-3}$. This
gives Equation (\ref{dist}).
\begin{equation}\label{dist}
s=\sqrt{\frac{4d^2}{3}}
\end{equation}
Where $s$ is the spacing of the 3-dimensional grid of sample points,
$1.53959*10^{-4}$. This equals roughly $4.57*10^{10}$ sample points
to check, another reason for choosing the simpler $L^1$-norm.
 The necessary calculations are easily performed by a
low-end desktop computer in less than 12 hours. Since the distance
from the weight position to the attractor is steadily diminishing,
it follows that the weight position will, given enough consecutive
inputs, come close enough to the attractor to reach ordered space.

\begin{figure}%
\centering \subfigure[Initial state.]{
          \label{sequence1}
          \includegraphics[width= 100pt]{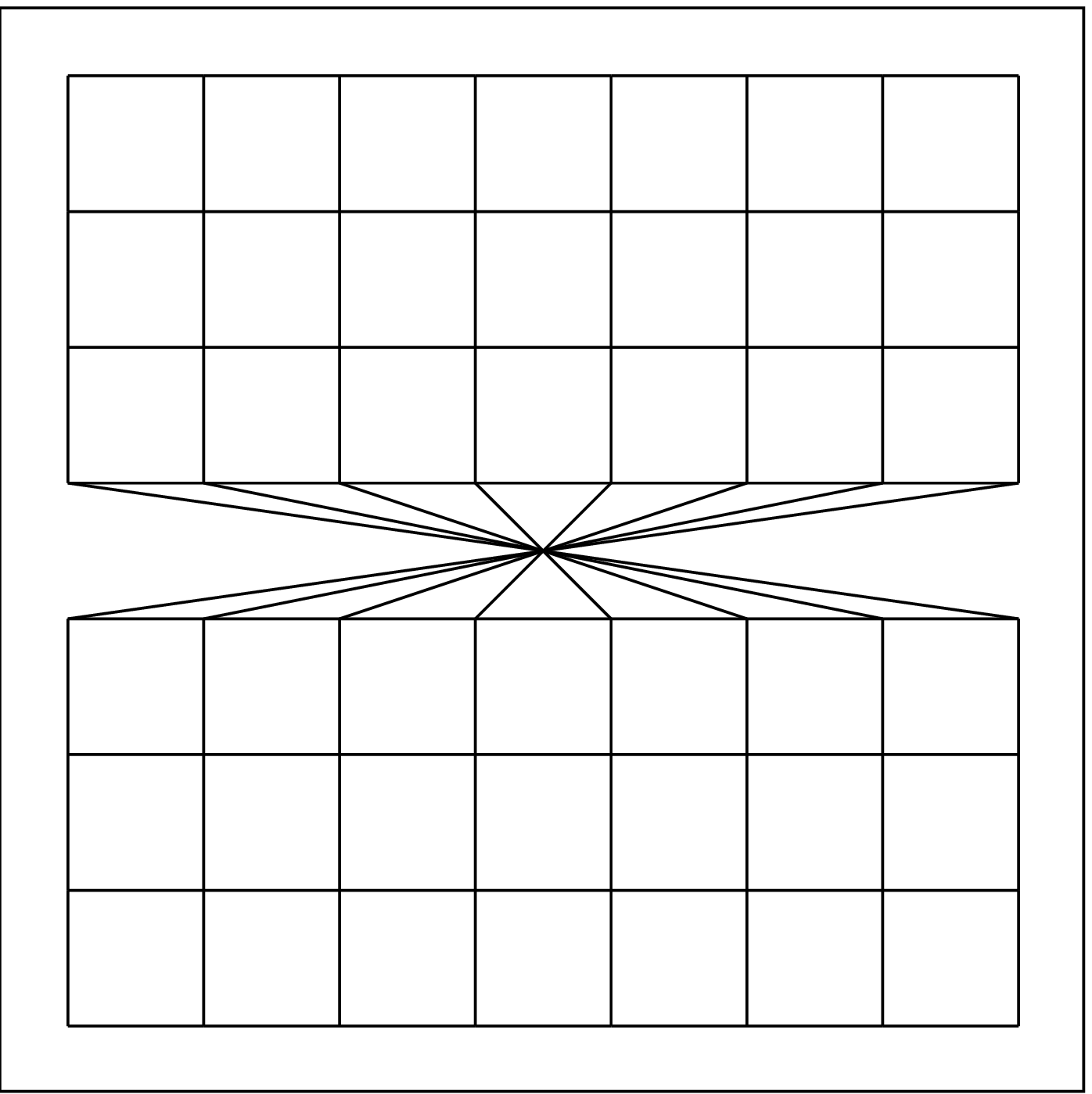}}%
          \hspace{6mm}
\subfigure[After 400 inputs.]{
          \label{sequence2}
          \includegraphics[width= 100pt]{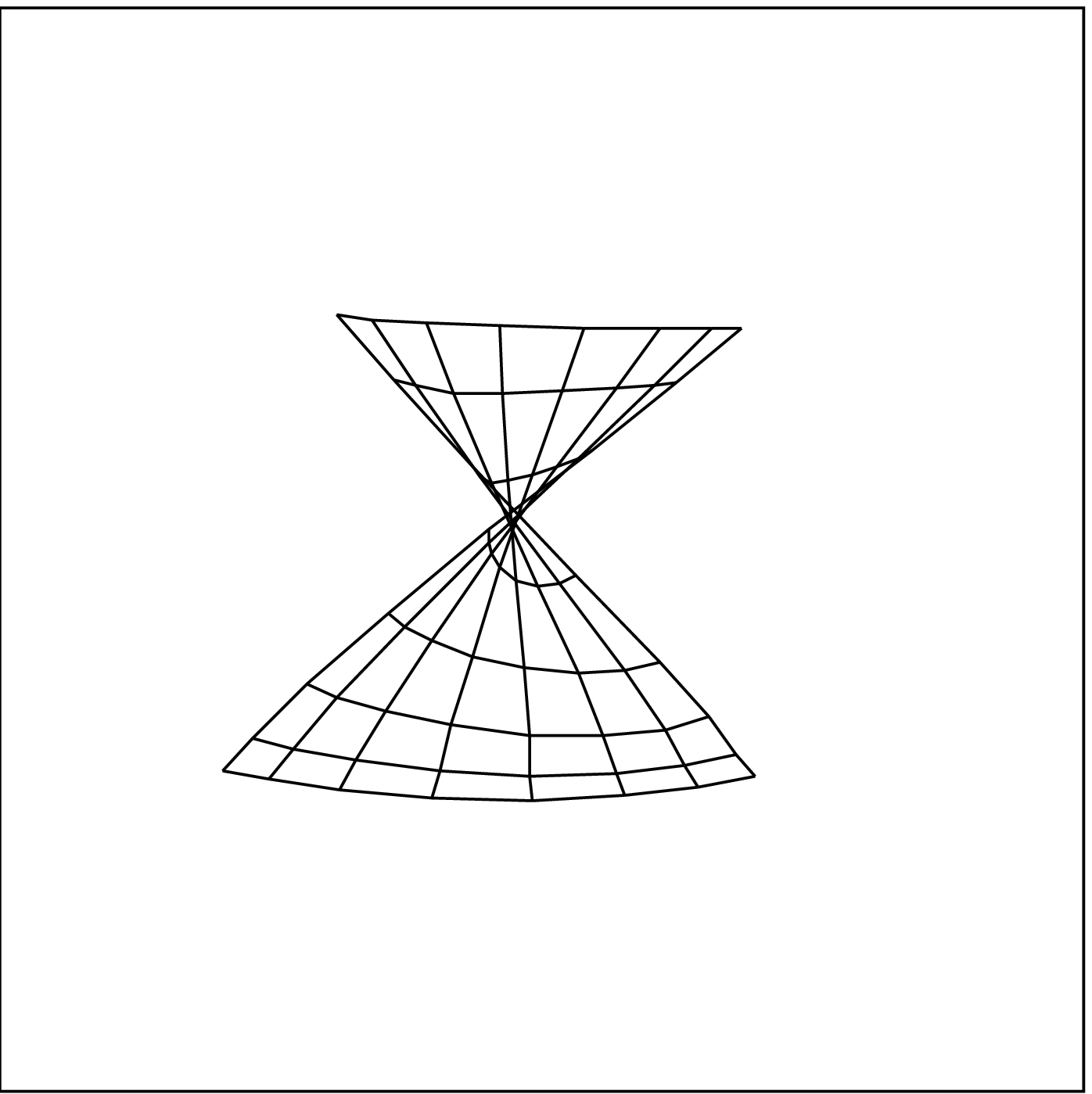}}\\
\subfigure[After 480 inputs.]{
          \label{sequence3}
          \includegraphics[width= 100pt]{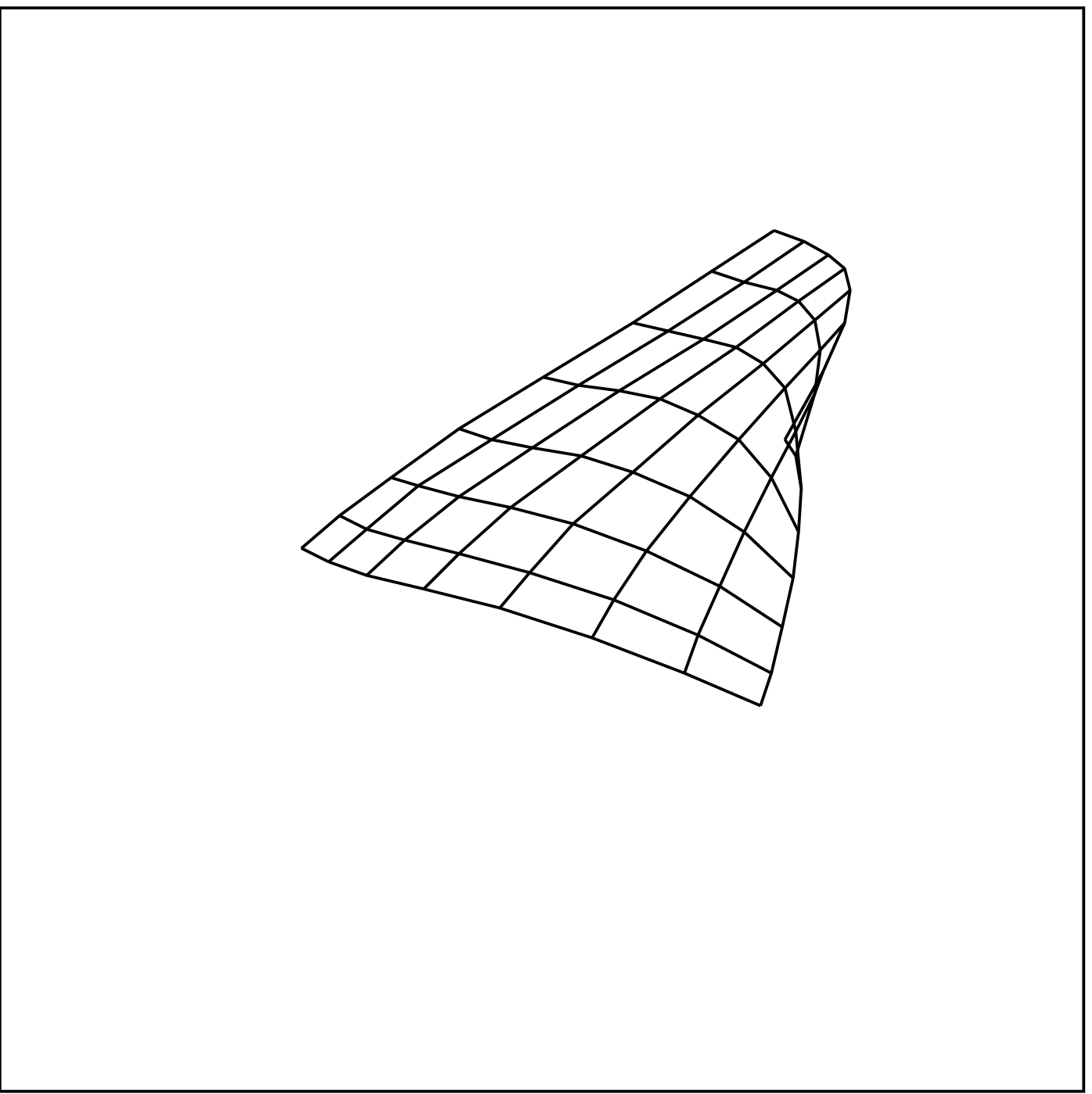}}%
          \hspace{6mm}
\subfigure[After 650 inputs.]{
          \label{sequence4}
          \includegraphics[width= 100pt]{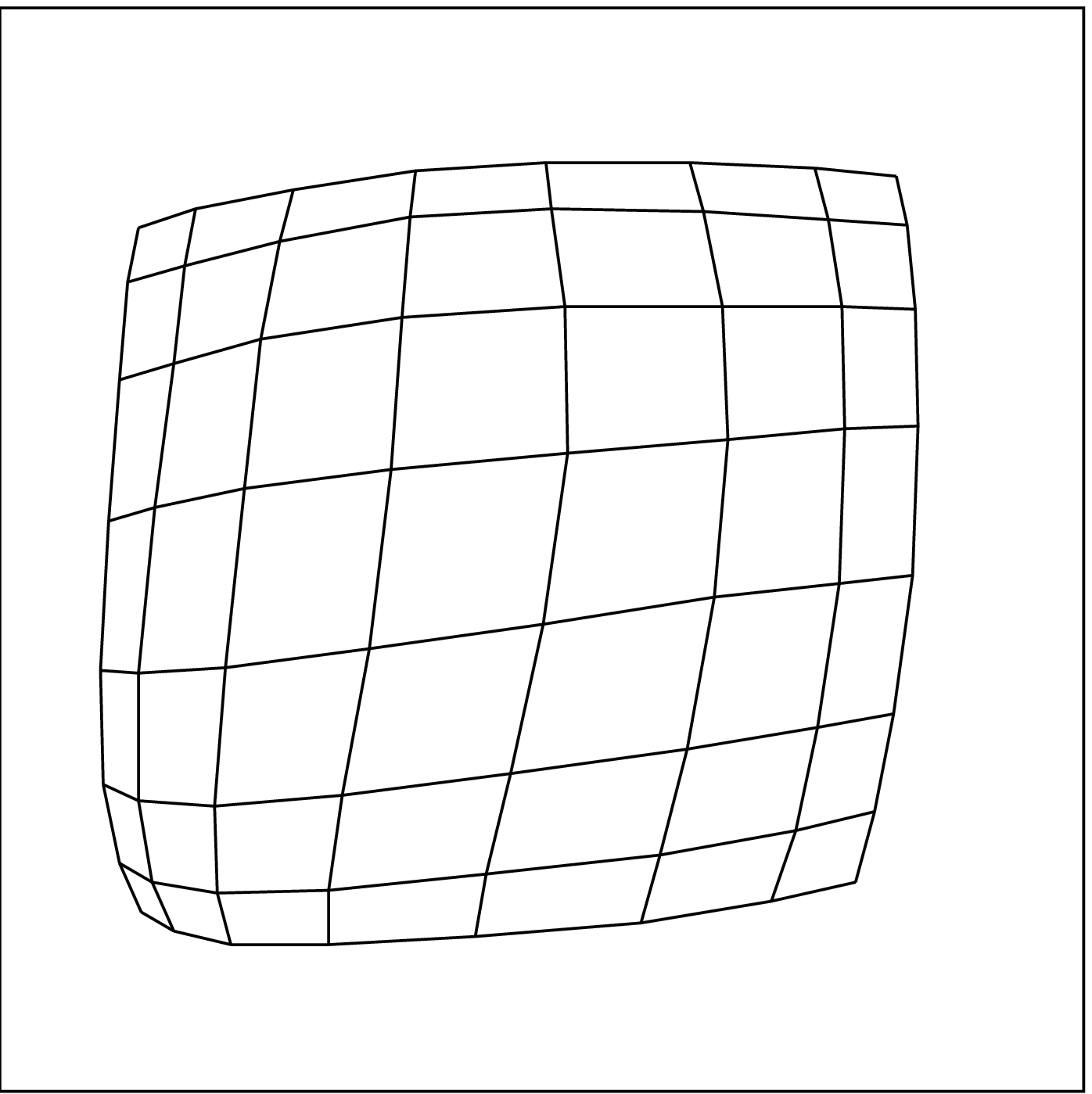}}\\
          \label{sequence}
          \caption{Evolution of the weight positions of a 64-node 2D PLSOM initialised
          to a difficult position. Neighbourhood size is 11, minimum neighbourhood size is 0.
    To simulate what will happen if this configuration appears late in training, we force an $r$ value of 0.65.}
\end{figure}

Whether this proof is extensible to networks with more than 3 nodes
and more than 1 dimensional input is at this point uncertain, but
the image sequence in (Figures \ref{sequence1}-\ref{sequence4})
certainly suggests the possibility.\\
Kohonen \cite{kohonen:1998} mentions a proof (see
\cite{cottrell:1998,cottrell:1987}) of ordering of a simplified SOM
based on the probability of an ordering input happening and an
infinite number of inputs. This proof in essence relies on the fact
that if there is a \emph{sequence of inputs} such that the map will
become ordered and one generates a sufficiently large number of
inputs the probability of encountering the ordering sequence of
inputs approach 1. The proof just presented here establishes that
for any configuration, the expected update is in the direction of
ordering for \emph{any single input}. It also shows the existence of
an ordered attractor for the dynamical system without having to
satisfy the Robbins-Monro \cite{robbins:1951} condition.

\textbf{Conjecture:} Any 1-dimensional PLSOM where only the
immediate neighbours of the winning node are updated can be seen as
a chain of 3-node networks, where each subnetwork is guaranteed to
become ordered, therefore the whole network will become ordered.
This is similar to the proof given in \cite{kohonen:1998} for the
SOM, albeit the authors are not confident enough that it also
applies to the PLSOM to posit it as more than a conjecture.
\section*{Acknowledgment}
The authors would like to thank Dr. Frederic Maire of the Smart
Devices Lab, Queensland University of Technology and Dr. Gordon
Wyeth of the division of Complex and Intelligent Systems, University
of Queensland for their valuable input.

\bibliographystyle{elsart-harv}\bibliography{TNN04-P288}

\begin{thebibliography}{46}
\expandafter\ifx\csname natexlab\endcsname\relax\def\natexlab#1{#1}\fi
\expandafter\ifx\csname url\endcsname\relax
  \def\url#1{\texttt{#1}}\fi
\expandafter\ifx\csname urlprefix\endcsname\relax\def\urlprefix{URL }\fi

\bibitem[{Berglund(2004)}]{berglund:2004}
Berglund, E., 2004. {PLSOM-controlled 6 degree of freedom robotic arm playing
  chess}. \\http://www.itee.uq.edu.au/\~{ }berglund/robochess.mov.

\bibitem[{Berglund and Sitte(2003)}]{berglund:2003}
Berglund, E., Sitte, J., 2003. {The Parameter-Less SOM algorithm}. In: ANZIIS
  2003. pp. 159--164.

\bibitem[{Bishop et~al.(1997)Bishop, Svens\'{e}n, and Williams}]{gtm}
Bishop, C.~M., Svens\'{e}n, M., Williams, C. K.~I., 1997. {GTM: A Principled
  Alternative to the Self-Organizing Map}. Advances in Neural Information
  Processing Systems 1~(9), 354--360.

\bibitem[{Bishop et~al.(1998)Bishop, Svens\'{e}n, and Williams}]{gtm2}
Bishop, C.~M., Svens\'{e}n, M., Williams, C. K.~I., 1998. {GTM: The Generative
  Topographic Mapping}. Neural Computation 10~(1), 215--235.

\bibitem[{Campbell et~al.(2005)Campbell, Berglund, and Streit}]{campbell:2005}
Campbell, A., Berglund, E., Streit, A., 2005. {Graphics Hardware Implementation
  of the Parameter-Less Self-Organising Map}. In: {Proc. of IDEAL'05}. pp.
  343--350.

\bibitem[{Cole et~al.(1990)Cole, Muthusamy, and Fanty}]{cole:1990}
Cole, R.~A., Muthusamy, Y.~K., Fanty, M., 1990. {The ISOLET Spoken Letter
  Database}. Tech. Rep. 90-004, Oregon Graduate Institute, Beaverton, Oregon,
  USA.

\bibitem[{Cottrell and Fort(1987)}]{cottrell:1987}
Cottrell, M., Fort, J., 1987. {Etude d'un algorithme d'auto-organisation}.
  {Annales de l'Institute Henri Poincare (B) Probabilit\'{e}s et statistiques}
  23~(1), 1--20.

\bibitem[{Cottrell et~al.(1998)Cottrell, Fort, and Pag{\`e}s}]{cottrell:1998}
Cottrell, M., Fort, J.~C., Pag{\`e}s, G., Nov 1998. {Theoretical aspects of the
  SOM algorithm}. Neurocomputing 21, 119--138.

\bibitem[{Dietterich and Bakiri(1991)}]{dietterich:1991}
Dietterich, T.~G., Bakiri, G., 1991. {Error-correcting output codes: A general
  method for improving multiclass inductive learning programs.} In: {Proc. of
  AAAI-91}. pp. 572--577.

\bibitem[{Fanty and Cole(1990)}]{fanty:1990}
Fanty, M., Cole, R., 1990. {Spoken letter recognition}. In: {Proc. of the 1990
  conference on Advances in neural information processing systems 3}. Morgan
  Kaufmann, pp. 220--226.

\bibitem[{Fix and Hodges(1951)}]{fix:1952}
Fix, E., Hodges, J., 1951. {Discriminatory analysis, non-parametric
  discrimination}. Tech. Rep.~4, USAF School of Aviation Medicine, Randolf
  Field, Texas, USA.

\bibitem[{Fritzke(1994)}]{fritzke:1994}
Fritzke, B., 1994. {Growing cell structures - a self-organizing network for
  unsupervised and supervised learning.} Neural Networks 7~(9), 1441--1460.

\bibitem[{Fritzke(1995)}]{fritzke:1995}
Fritzke, B., 1995. {A growing neural gas network learns topologies}. MIT Press,
  pp. 625--632.

\bibitem[{G{\"o}ppert and Rosenstiel(1996)}]{goppert:1996}
G{\"o}ppert, J., Rosenstiel, W., 1996. {Varying cooperation in SOM for improved
  function approximation}. In: IEEE Int. Conf. on Neural Networks. Vol.~1. pp.
  1--6.

\bibitem[{{H. Kitano and H. G. Okuno and K. Nakadai and T. Matsui and K. Hidai
  and T. Lourens}(2002)}]{kitano:2002}
{H. Kitano and H. G. Okuno and K. Nakadai and T. Matsui and K. Hidai and T.
  Lourens}, 2002. {SIG, The Humanoid}.
  \\http://www.symbio.jst.go.jp/symbio/SIG/.

\bibitem[{Haese(1999)}]{haese:1999}
Haese, K., 1999. {Kalman Filter Implementation of Self-Organizing Feature
  Maps.} Neural Computation 11~(5), 1211--1233.

\bibitem[{Haese and Goodhill(2001)}]{haese:2001}
Haese, K., Goodhill, G.~J., 2001. {Auto-SOM: Recursive Parameter Estimation for
  Guidance of Self-Organizing Feature Maps.} Neural Computation 13~(3),
  595--619.

\bibitem[{Hartmann(1999)}]{hartmann:1999}
Hartmann, W.~M., 1999. {How we localize sound}. Physics Today 1~(November),
  23--29.

\bibitem[{Huang et~al.(1995)Huang, Ohnishi, and Sugie}]{huang:1995}
Huang, J., Ohnishi, N., Sugie, N., 1995. {A Biometric System for Localization
  and Separation of Multiple Sound Sources}. IEEE Trans. on Instrumentation and
  Measurement 44~(3), 733--738.

\bibitem[{Huang et~al.(1997)Huang, Ohnishi, and Sugie}]{huang:1997}
Huang, J., Ohnishi, N., Sugie, N., 1997. {Building ears for robots: Sound
  localization and separation}. Artificial Life and Robotics 1~(4), 157--163.

\bibitem[{Iglesias and Barro(1999)}]{iglesias:1999}
Iglesias, R., Barro, S., 1999. {SOAN: Self Organizing with Adaptive
  Neighborhood Neural Network.} In: IWANN (1). pp. 591--600.

\bibitem[{Kaas(1991)}]{neuralplast}
Kaas, J.~H., Mar 1991. {Plasticity of Sensory and Motor Maps in Adult Mammals}.
  Annual Review of Neuroscience 14, 137--167.

\bibitem[{Keeratipranon and Maire(2005)}]{keeratipranon:2005}
Keeratipranon, N., Maire, F., 2005. {Bearing Similarity Measures for
  Self-Organizing Feature Maps}. In: {Proc. of IDEAL'05}. pp. 286--293.

\bibitem[{King and Carlile(1995)}]{king:1995}
King, A., Carlile, S., 1995. {Neural Coding for Auditory Space}. The Cognitive
  Neurosciences. The Mit Press, London, England.

\bibitem[{Kohonen(1989)}]{kohonen:1998}
Kohonen, T., 1989. {Self-Organization and Associative Memory}, 3rd Edition.
  Springer-Verlag.

\bibitem[{Kohonen(1990)}]{kohonen:1990a}
Kohonen, T., 9 1990. {The self-organizing map}. Proc. of the IEEE 78~(9),
  1464--1480.

\bibitem[{Kwok and Smith(2004)}]{kwok:2004}
Kwok, T., Smith, K.~A., 2004. {A Noisy Self-Organizing Neural Network With
  Bifurcation Dynamics for Combinatorial Optimization}. {IEEE Trans. on Neural
  Networks} 15~(1), 84--98.

\bibitem[{Lang and Warwick(2002)}]{lang:2002}
Lang, R., Warwick, K., 2002. {The plastic self organising map}. In: Proc. of
  the 2002 Int. Joint Conf. on Neural Networks. Vol.~1. pp. 727--732.

\bibitem[{Lourens et~al.(2000)Lourens, Nakadai, Okuno, and
  Kitano}]{lourens:2000}
Lourens, T., Nakadai, K., Okuno, H.~G., Kitano, H., 2000. {Humanoid Active
  Audition System}. In: IEEE-RAS Int. Conf. on Humanoid Robots.

\bibitem[{Nakadai et~al.(2000)Nakadai, Lourens, Okuno, and
  Kitano}]{nakadai:2000a}
Nakadai, K., Lourens, T., Okuno, H.~G., Kitano, H., 2000. {Active Audition for
  Humanoid}. In: AAAI-2000. pp. 832--839.

\bibitem[{Nakadai et~al.(2002)Nakadai, Okuno, and Kitano}]{nakadai02realtime}
Nakadai, K., Okuno, H., Kitano, H., 2002. {Realtime sound source localization
  and separation for robot audition}. In: {Proc. IEEE Int. Conf. on Spoken
  Language Processing}. pp. 193--196.

\bibitem[{Nakadai et~al.(2003)Nakadai, Okuno, and Kitano}]{nakadai:2003}
Nakadai, K., Okuno, H.~G., Kitano, H., Sept. 2003. {Robot recognizes three
  simultaneous speech by active audition}. In: ICRA '03. Vol.~1. pp. 398--405.

\bibitem[{Nakashima et~al.(2002)Nakashima, Mukai, and
  Ohnishi}]{nakashima:2002a}
Nakashima, H., Mukai, T., Ohnishi, N., November 2002. {Self-Organization of a
  sound source localization robot by perceptual cycle}. In: ICONIP'02. Vol.~2.
  pp. 834--838.

\bibitem[{Nakatani et~al.(1994)Nakatani, Okuno, and Kawabata}]{nakatani:1994}
Nakatani, T., Okuno, H.~G., Kawabata, T., 1994. {Auditory Stream Segregation in
  Auditory Scene Analysis with a Multi-Agent System.} In: AAAI-94. pp.
  100--107.

\bibitem[{Nandy and Ben-Arie(1995)}]{nandy:1995}
Nandy, D., Ben-Arie, J., 1995. {Auditory Localization Using Spectral
  Information}. Academic Press.

\bibitem[{Rabinkin et~al.(1996)Rabinkin, Renomeron, Dahl, French, Flanagan, and
  Bianchi}]{rabinkin:1996}
Rabinkin, D., Renomeron, R., Dahl, A., French, J., Flanagan, J., Bianchi, M.,
  1996. {A DSP Implementation of Source Location Using Microphone Arrays}.
  Proc. of the SPIE 2846, 88--99.

\bibitem[{Ritter et~al.(1992)Ritter, Martinetz, and Schulten}]{ritter:1992}
Ritter, H., Martinetz, T., Schulten, K., 1992. {Neural Computation and
  Self-Organizing Maps - An Introduction}. Addison-Wesley publishing company.

\bibitem[{Robbins and Muonro(1951)}]{robbins:1951}
Robbins, H., Muonro, S., 1951. {A stochastic approximation method}. The Annals
  of Mathematical Statistics 22, 400--407.

\bibitem[{Rucci et~al.(1999)Rucci, Edelman, and Wray}]{rucci:1999}
Rucci, M., Edelman, G., Wray, J., 1999. {Adaptation of Orienting Behavior: From
  the Barn Owl to a Robotic System}. {IEEE Trans. on Robotics and Automation}
  15~(1), 96--110.

\bibitem[{Shah-Hosseini and Safabakhsh(2000)}]{shah-hosseini:2000}
Shah-Hosseini, H., Safabakhsh, R., 2000. {TASOM: The Time Adaptive
  Self-Organizing Map.} In: ITCC. IEEE Computer Society, pp. 422--427.

\bibitem[{Shah-Hosseini and Safabakhsh(2003)}]{shah-hosseini:2003}
Shah-Hosseini, H., Safabakhsh, R., 2003. {TASOM: a new time adaptive
  self-organizing map.} IEEE Trans. on Systems, Man, and Cybernetics, Part B
  33~(2), 271--282.

\bibitem[{Starzyk et~al.({2005})Starzyk, Zhu, and Liu}]{starzyk:2005}
Starzyk, J.~A., Zhu, Z., Liu, T.-H., {2005}. {Self-Organizing Learning Array}.
  {IEEE Trans. on Neural Networks} 16~(2), 355--363.

\bibitem[{Sutton(1992)}]{sutton:1992}
Sutton, R.~S. (Ed.), 1992. {Reinforcement Learning}. Kluwer Academic
  Publishers.

\bibitem[{Vellido et~al.(2003)Vellido, El-Deredy, and Lisboa}]{vellido:2003}
Vellido, A., El-Deredy, W., Lisboa, P. J.~G., July 2003. {Selective smoothing
  of the generative topographic mapping}. IEEE Trans. on Neural Networks
  14~(4), 847--852.

\bibitem[{Wang and Chen(1991)}]{wang:1991}
Wang, L.-C., Chen, C., Aug 1991. {A combined optimization method for solving
  the inverse kinematics problems of mechanical manipulators}. IEEE Trans. on
  Robotics and Automation 7~(4), 489--499.

\bibitem[{Zwiers et~al.(2001)Zwiers, Ostal, and Cruysberg}]{zwiers:2001}
Zwiers, M.~P., Ostal, A. J.~V., Cruysberg, J. R.~M., 2001. {Two-dimensional
  sound-localization behavior of early-blind humans}. Experimental Brain
  Research 140, 206--222.

\end{thebibliography}

\end{document}